\documentclass[letterpaper, 10 pt, conference]{ieeeconf}
\IEEEoverridecommandlockouts                              % This command is only needed if
\usepackage{times}
\usepackage{epsfig}
\usepackage{graphicx}
\usepackage{amsmath}
\usepackage{amssymb}
\usepackage{graphicx}
\usepackage{amsmath}
\usepackage{booktabs}
\usepackage{algorithm}
\usepackage{algorithmic}
\usepackage{comment}
\usepackage{graphicx}
\usepackage{amsmath,amssymb} % define this before the line numbering.
\usepackage{color}
\usepackage{booktabs}

\usepackage[pagebackref=true,breaklinks=true,letterpaper=true,colorlinks,bookmarks=false]{hyperref}

\author{Jinchang Zhang*, Praveen Kumar Reddy*, Xue-Iuan Wong, Yiannis Aloimonos, Guoyu Lu
\thanks{* indicates equal contribution. Jinchang Zhang, Praveen Kumar Reddy, and Guoyu Lu are with the Intelligent Vision and Sensing (IVS) Lab at the University of Georgia
        {\tt\small guoyulu62@gmail.com}, Xue-Iuan Wong is with Ford Motor Company. Yiannis Aloimonos is with the University of Maryland.}%
}

\begin{document}

\title{Embodiment: Self-Supervised Depth Estimation Based on Camera Models}

\maketitle

\begin{abstract}
Depth estimationn is a critical topic for robotics and vision-related tasks. In monocular depth estimation, in comparison with supervised learning that requires expensive ground truth labeling, self-supervised methods possess great potential due to no labeling cost. However, self-supervised learning still has a large gap with supervised learning in 3D reconstruction and depth estimation performance. Meanwhile, scaling is also a major issue for monocular unsupervised depth estimation, which commonly still needs ground truth scale from GPS, LiDAR, or existing maps to correct. 
In the era of deep learning, existing methods primarily rely on exploring image relationships to train unsupervised neural networks, while the physical properties of the camera itself—such as intrinsics and extrinsics—are often overlooked. These physical properties are not just mathematical parameters; they are embodiments of the camera's interaction with the physical world. By embedding these physical properties into the deep learning model, we can calculate depth priors for ground regions and regions connected to the ground based on physical principles, providing free supervision signals without the need for additional sensors. This approach is not only easy to implement but also enhances the effects of all unsupervised methods by embedding the camera's physical properties into the model, thereby achieving an embodied understanding of the real world.
\end{abstract}

\section{INTRODUCTION}
\label{sec:intro}
Monocular depth estimation serves as a cornerstone in robotics \cite{hazirbas2017fusenet}, 3D mapping \cite{li2023bevdepth}, camera localization \cite{localizeiccv, indoor}, and augmented reality \cite{tang2022perception}. This process aims to derive depth from a single RGB image, an inherently challenging task due to scale ambiguity, where one 2D image might represent numerous possible 3D scenes. Convolutional neural networks have significantly advanced this area \cite{he2016deep}, though most cutting-edge methods rely on supervised training \cite{eigen2014depth,bhat2021adabins}, necessitating sparse depth ground truth from instruments like LiDAR. The high cost and labor of data collection and labeling constrain the data scale for supervised approaches \cite{bhat2021adabins}. To avoid the need for depth labeling, there has been a shift towards self-supervised frameworks, employing regression modules for pixel-wise depth estimation and photometric consistency loss for model training \cite{godard2019digging}. Despite these efforts, self-supervised learning for monocular depth estimation still faces significant accuracy challenges, often misestimating objects' 3D structures as either too distant or too close due to the indirect nature of photometric and cross-frame consistency constraints.

Meanwhile, due to the convenience brought by deep neural networks, extensive information from the sensors themselves has been ignored. This paper introduces a method that leverages camera model parameters (both intrinsic and extrinsic) to accurately calculate depth information, 
thereby embedding the camera model and its physical characteristics into the deep learning model. This approach goes beyond mere mathematical computation; it embodies the interaction between the camera and the physical world. By integrating these physical characteristics into the deep learning model,
we can accurately determine the depth for much of the scene, facilitating neural network training without the need for explicit ground truth data. The method also incorporates image semantics to calculate the ground plane's depth, allowing for the estimation of the depth of objects on the ground, such as buildings and vehicles. Utilizing this physics-based supervision approach, the framework not only enhances the performance of unsupervised networks but also achieves an embodied understanding of the real world, providing strong support for detailed 3D structure modeling. Importantly, our algorithm serves as a valuable extension for any unsupervised depth estimation effort.

\begin{figure*}[t]
\begin{center}
%\framebox[4.0in]{$\;$}
\includegraphics[width=16cm, height=7cm]{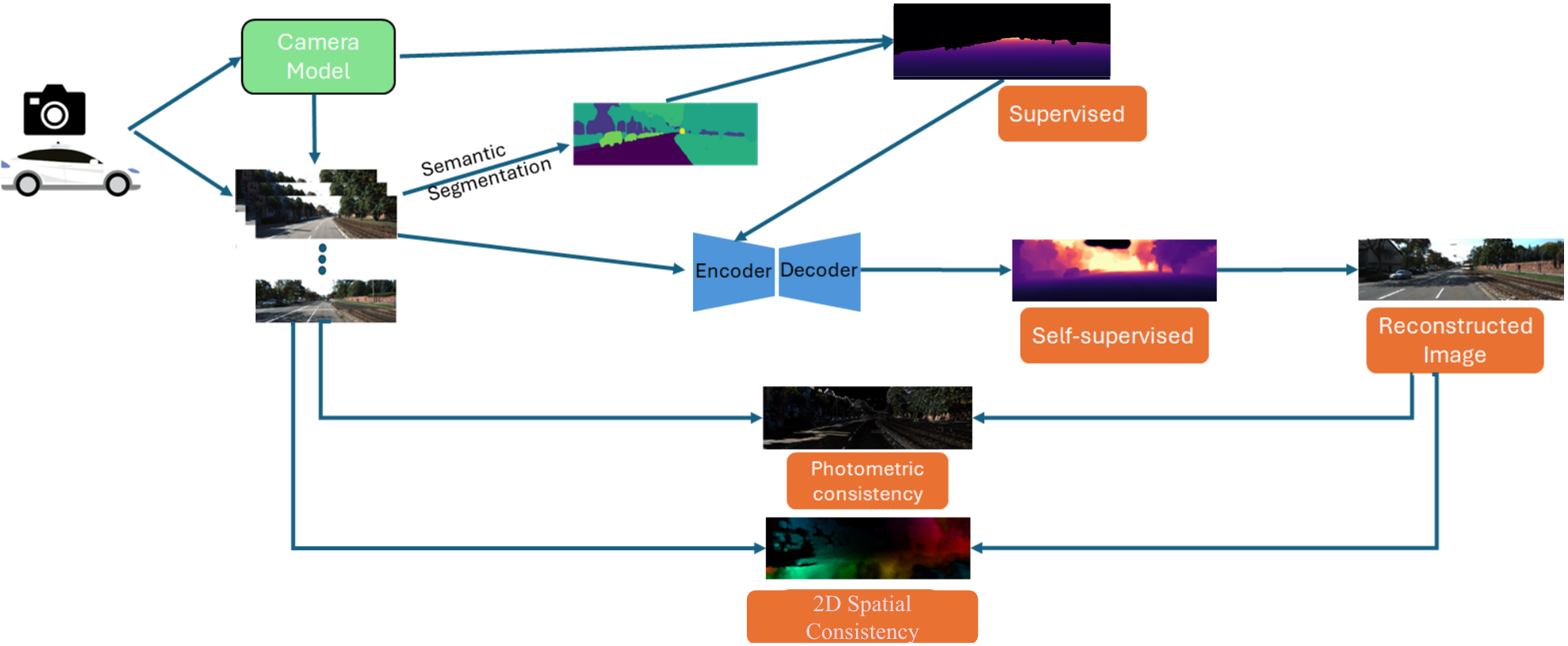}
\end{center}
\vspace{-4mm}
\caption{The framework for unsupervised 3D scene reconstruction neural network based on physics depth calculated from camera models. We first calculate the physical depth of flat ground areas in the input image using the camera model and semantic segmentation results. This physical depth serves as a label for supervised learning, providing a foundation for initial depth estimation. In the first stage, we train the depth estimation network with these labels. In the subsequent self-supervised stage, we introduce photometric and 2D spatial losses, which optimize depth estimation based on image characteristics without relying on depth labels. }
\vspace{-6 mm}
\label{overallframework}
\end{figure*}

In summary, our main contributions include the following aspects:
1. We propose a novel mechanism that leverages the physical model parameters of the camera to calculate the depth information for a large portion of the scene, embedding the camera model and its physical properties into the deep learning model to supervise the depth estimation network. We refer to this depth information, derived from the camera’s physical model, as physics depth.
2. To address the uncertain scale issue in unsupervised monocular depth estimation, our approach provides an absolute scale instead of just a relative scale.
3. We designed a neural network training framework to effectively integrate physics depth supervision with unsupervised methods, specifically targeting the physics depth calculated from the camera model.
The framework for physics depth computation and training with the self-supervised network is shown in Fig. \ref{overallframework}.

\section{Relate Work}

% \subsection{Supervised Depth Estimation}
% In the field of monocular 3D reconstruction and depth estimation based on neural networks, \cite{eigen2014depth} stands as a cornerstone contribution. Their pioneering work introduced a coarse-to-fine convolution neural network alongside a scale-invariant loss function. Subsequently, monocular depth estimation and 3D reconstruction have garnered considerable attention, focusing on enhancing performance through increasingly sophisticated architectures and loss functions \cite{laina2016deeper,lee2018single,liu2015learning,miangoleh2021boosting}. Some scholars have redefined the challenge as an ordinal regression task. Currently, supervised learning in monocular depth estimation and 3D reconstruction is primarily divided into two methodologies: a pixel-wise regression approach \cite{eigen2014depth,zhao2021transformer,ranftl2021vision} and a pixel-wise classification framework \cite{fu2018deep,diaz2019soft}. While regression methods facilitate the prediction of continuous depths, they pose optimization challenges. In contrast, classification methods enable discrete depth predictions and are comparatively more straightforward to optimize.

\subsection{Self-Supervised Depth Estimation}
Self-supervised depth estimation from monocular videos or stereo image pairs is gaining prominence, particularly due to the challenges in obtaining accurate ground truth. \cite{zhou2017unsupervised} spearheaded a self-supervised framework by jointly training depth and pose networks based on image reconstruction loss. \cite{godard2019digging} further advanced this field by introducing a minimum reprojection loss and auto-masking loss, setting a new benchmark. \cite{guizilini20203d} and \cite{chawla2021multimodal} integrated real-time data sources such as GPS or camera velocity to address the scale ambiguity inherent in monocular Structure-from-Motion (SfM) methods. These self-supervised models hinge on the photometric consistency of the reprojection. In stereo training contexts, models use synchronous stereo image pairs to predict disparity \cite{scharstein2002taxonomy}, which inversely relates to depth. Since the relative camera pose is known in stereo setups, the primary task of these models is disparity prediction. \cite{garg2016unsupervised} pioneered this approach, training a self-supervised monodepth model using stereo pairs and a photometric consistency loss. This methodology was further refined by \cite{godard2017unsupervised} with additional constraints like left-right consistency and bundle adjusted pose graph \cite{lu2023deep}. Moreover, \cite{garg2020wasserstein} extended it to predict continuous disparity. Stereo views naturally offer an absolute depth scale, whereas current monocular self-supervised models only predict relative depths, needing ground truth for scale calibration. Utilizing physics depth data from ground surfaces and road-connecting areas can improve these unsupervised models to predict absolute depths, enhancing accuracy for datasets such as KITTI.

\subsection{Geometric Priors}
Geometric priors are becoming crucial for monocular depth estimation, evolving beyond the optimization-focused traditional multi-view stereo methods noted by \cite{gallup2010piecewise}. In self-supervised learning, multi-view geometry is essential, enabling image warping from source to target viewpoints to generate reprojection errors. These errors act as the loss function for depth estimation networks \cite{godard2019digging}. Moreover, geometric consistency, especially in comparing point clouds, is gaining attention as a valuable complement to photometric consistency \cite{mahjourian2018unsupervised}. The surface normal constraint, highlighted in works by \cite{kusupati2020normal} and \cite{long2021adaptive}, is a key geometric prior ensuring the alignment of normal vectors from both estimated and actual depth data. However, this approach can lead to inaccuracies in depth estimations, especially in areas with high curvature, underscoring the limitations of relying solely on planarity assumptions.

%-------------------------------------------------------------------------

\section{Embodiment Physics Depth}
\label{sec:formatting}
\subsection{Physics depth for Full Field of view}
We introduce a novel monocular depth estimation technique called "physics depth".
This method leverages the camera's intrinsic and extrinsic parameters, as well as semantic segmentation, to calculate absolute depth, embedding the physical characteristics of the camera model into the deep learning model.
This approach derives depth from basic physical principles, assuming the camera captures initially flat surfaces. We refine depth estimates by identifying truly flat areas through semantic segmentation, extending depth to adjacent areas, and filling gaps with inpainting. Our method, tested on KITTI, CityScape, and Make3D datasets, achieves accuracy comparable to LiDAR, especially for close, flat surfaces.
Our model is based on the pinhole camera model, which is widely used in practical applications due to its minimal distortion, making it an ideal reference point for achieving embodiment.
Although designed with the pinhole model in mind, our method can be adapted for different camera types by adjusting for each camera's unique characteristics. For every image pixel, we compute a unit vector $\hat{r}$ representing the camera ray's direction in the physical world, translating pixel positions into directional vectors that indicate the camera's viewpoint in real space.
\vspace{-2mm}
\begin{equation}
\centering
\vspace{-2mm}
\label{unitvector}
\begin{array}{l}
\hat{r} = \frac{{\left[ {u,v,f} \right]}}{{\sqrt {{u^2} + {v^2} + {f^2}} }}
\end{array}
\end{equation}

\noindent where ($u$, $v$) represents the coordinates of the pixel, with the origin of the coordinate system situated at the optical center (${{O_x}, {O_y}}$) of the image, commonly referred to as the principal point. Meanwhile, $f = ({{f_x} + {f_y}})/2$, where ${f_x}$ and ${f_y}$ denote the camera's focal length in the $x$ and $y$ directions.

To generate a physics depth scaled to dimensions different from those of the original RGB image, the parameters of the unit vector $\hat{r}$ must be adjusted accordingly. Suppose \( W_{\text{org}} \) and \( H_{\text{org}} \) are the width and height of the original RGB image, and \( W_{\text{new}} \) and \( H_{\text{new}} \) are the desired width and height for the physics depth. Let \( S_{\text{width}} \) and \( S_{\text{height}} \) represent the scaling factors for the width and height, respectively, where \( S_{\text{width}} = \frac{W_{\text{new}}}{W_{\text{org}}} \) and \( S_{\text{height}} = \frac{H_{\text{new}}}{H_{\text{org}}} \). The scale-adjusted pixel coordinates ($u', v'$) are given by ($S_{\text{width}} \times u, S_{\text{height}} \times v$). The scale-adjusted optical center coordinates ($O_x', O_y'$) are ($S_{\text{width}} \times O_x, S_{\text{height}} \times O_y$), and the scale-adjusted focal lengths ($f_x', f_y'$) are ($S_{\text{width}} \times f_x, S_{\text{height}} \times f_y$), as determined by perspective projection. The scale-adjusted unit vector $\hat{r'}$ can be derived using the below equation by updating the parameters in Eq. \ref{unitvector}:

\begin{equation}
\begin{aligned}
R_{roll} &= \left[ \begin{array}{ccc}
1 & 0 & 0 \\
0 & \cos(\text{roll}) & \sin(\text{roll}) \\
0 & -\sin(\text{roll}) & \cos(\text{roll})
\end{array} \right], \\
R_{pitch} &= \left[ \begin{array}{ccc}
\cos(\text{pitch}) & 0 & -\sin(\text{pitch}) \\
0 & 1 & 0 \\
\sin(\text{pitch}) & 0 & \cos(\text{pitch})
\end{array} \right], \\
R_{yaw} &= \left[ \begin{array}{ccc}
\cos(\text{yaw}) & \sin(\text{yaw}) & 0 \\
-\sin(\text{yaw}) & \cos(\text{yaw}) & 0 \\
0 & 0 & 1
\end{array} \right]
\end{aligned}
\end{equation}

\vspace{-2mm}
\begin{equation}
\centering
\begin{array}{l}
{R_c} = {R_{yaw}}*{R_{pitch}}*{R_{roll}}
\end{array}
\vspace{-2mm}
\end{equation}

Using $R_c$ we rotate the camera ray vector to align it with the ground coordinate system: 
$\hat{r_c} = {R_c}* \hat{r'}$

Since $\hat{r_c} ({r_{c,u}},{r_{c,v}},{r_{c,f}})$ is a unit vector, the 3D coordinates of the point, $P=(x_c, y_c, z_c)$, on the ground surface in camera's coordinate system can be determined by multiplying $r_c$ with the point-to-point distance ($d$) of the ground point from camera.
\vspace{-1.5mm}
\begin{equation}
\begin{aligned}
& \left[x_c, y_c\right]=d * \left[ {{r_{c,u}},{r_{c,v}}} \right]
\end{aligned}
\vspace{-2mm}
\end{equation}

\noindent When the height of the camera ($h$) is known from the camera's extrinsic parameters and assuming the camera coordinate system's y-axis is oriented downwards, then ${y_c}$ = $h$, and the point-to-point distance $d$ and ${x_c}$ can be calculated as shown:
\vspace{-3mm}
\begin{equation}
d = \frac{h}{{{r_{c,v}}}} ,   {x_c} = d * {r_{c, u}}
\vspace{-2mm}
\end{equation}

The projection of a three-dimensional point from the camera coordinate system \((x_c, y_c, z_c)\) to the two-dimensional image plane \((u, v)\), can be accurately represented using the following linear camera model equation:

\begin{equation}
\label{project1}
\setlength{\arraycolsep}{2pt}
{Z_c}\left[ \begin{array}{c}
u\\
v\\
1
\end{array} \right] = \left[ \begin{array}{ccc}
{{f_x'}} & 0 & {{O_x'}}\\
0 & {{f_y'}} & {{O_y'}}\\
0 & 0 & 1
\end{array} \right]\!\left[ \begin{array}{c}
{{x_c}}\\
{{y_c}}\\
{{z_c}}\\
\end{array} \right]
\end{equation}

where \( \mathbf{K} \) denotes the camera's intrinsic matrix:

\vspace{-1mm}
\begin{equation}
\setlength{\arraycolsep}{2pt}
{K} = \left[ \begin{array}{ccc}
{{f_x'}} & 0 & {{O_x'}}\\
0 & {{f_y'}} & {{O_y'}}\\
0 & 0 & 1
\end{array} \right]
\end{equation}

By substituting \(x_c\), \(y_c\) in Eq. \ref{project1}, we derive \(z_c\) for a given pixel \((u, v)\) that maps to a ground point. This process allows calculating depth and 3D coordinates in the camera coordinate system for ground surface pixels, given the camera's height. We tested our approach on the KITTI \cite{geiger2013vision} and Cityscapes \cite{cordts2016cityscapes} datasets, with results detailed in Section \ref{section:Physics Depth Demonstration}.

% \begin{figure}[t]
% \begin{center}
% %\framebox[4.0in]{$\;$}
% \includegraphics[width=8cm, height=3cm]{image/icp.png}
% \end{center}
% \vspace{-6mm}
% \caption{\textbf{The point cloud matching process:} For each timestep \( t \), an estimated depth \( D_t \) is used to generate a corresponding point cloud \( Q_t \). This point cloud is subsequently transformed by the estimated pose \( T_t \), yielding a predicted point cloud \( \hat{Q}_{t-1} \) for the preceding frame. Should the Iterative Closest Point (ICP) algorithm identify a more accurate registration between \( Q_{t-1} \) and \( \hat{Q}_{t-1} \), the pose estimation is refined accordingly with the correction \( T'_t \). Residuals \( r_t \), observed post-registration, indicate discrepancies in the depth \( D_t \). }
% \vspace{-4mm}
% \label{icp}
% \end{figure}
% \vspace{-2mm}
% \subsection{Extension of Physics Depth}

Our method improves depth estimation by closely aligning it with LiDAR data on flat surfaces like roads, providing more detail than standard sparse LiDAR data, as demonstrated on the KITTI and Cityscapes datasets. Initially targeting flat surfaces may risk overfitting, limiting versatility. To counter this, we expanded our physics depth approach to cover the entire image, including vertical structures like cars and buildings. This involves deriving depth by extending upwards from where flat and vertical surfaces meet, creating a comprehensive ground physics depth as detailed in Section \ref{section:Physics Depth Demonstration}.
We extended physics depth to vertical objects in contact with flat surfaces, such as vehicles, pedestrians, and buildings, by propagating depth values upward from their points of intersection with the ground, termed as Edge Extended Physics Depth. We assume these vertical entities have a consistent depth with the ground they touch, enabling us to infer their depth directly from the ground. This approach greatly improves the accuracy and consistency of depth estimation throughout the image. After vertically extending physics depth, we encountered objects with incomplete depth due to their limited contact with the ground. We used the Telea Inpainting Technique \cite{telea2004image} to fill these gaps, leveraging its fast and effective method based on the surrounding pixels' directional changes and geometric distances. For objects not touching the ground, we projected depth from nearby objects to ensure continuity. Additionally, we assigned the sky a depth 1.5 times the maximum of the inpainted depths, achieving a gap-free Dense Physics Depth. This approach primarily serves to create an improved depth prior, enhancing overall depth estimation accuracy.

The effectiveness of our method has been validated using the KITTI \cite{geiger2013vision} and Cityscapes \cite{cordts2016cityscapes} datasets, with results showing a close alignment in accuracy with LiDAR-derived depth measurements, especially for proximal flat surfaces.

\section{Interaction Between Physics-Depth Supervision and Self-Supervision}
\subsection{Network Architecture}
Our research addresses the data scarcity in physics depth, which is typically limited to only portions of an image and, by itself, inadequate for self-supervised learning. 
By embedding the physical characteristics of the camera model into the deep learning model, we utilize physics depth as an embodied prior, enhancing depth estimation. Unlike traditional self-supervised models that typically start with random depth values, our model can more accurately refine the depth estimation of ground surfaces and surrounding areas.
This approach significantly improves efficiency in correcting depth inaccuracies through self-supervised training. Our model uniquely combines RGB and physics depth, adding valuable depth insights. During the supervised phase, we assess the confidence in physics depth to focus learning on the reliable areas. For self-supervised learning, we advance the model by integrating geometric consistency with photometric consistency, leading to more precise depth estimates.

\subsection{Physics-Depth Supervision }
In this study, we calculate the physical depth of ground areas in each image using the camera model and employ it as an initial guide during the depth network training phase to enhance the model's understanding of depth across various regions. These physical depth data serve as labels for supervised learning, significantly improving the model's comprehension and prediction of ground area depths.
Specifically, we first utilize the camera's intrinsic and extrinsic parameters, along with the semantic segmentation results of the images, to accurately calculate the physical depth of ground areas. These physical depth data provide initial guidance to the model, equipping it with a fundamental understanding of spatial depth. This not only ensures the model has reliable depth information at the initial training stage but also effectively reduces dependence on randomly initialized depths, thereby preventing substantial error propagation during early training.
During training, the physical depth data act as supervisory signals, enabling the model to better learn the geometric structure of ground areas. This approach allows the model to more accurately capture the actual depth information of the ground and extend this understanding to the entire scene. The reliability of physical depth ensures the model performs more robustly and accurately when handling areas with similar geometric characteristics.
Furthermore, physical depth serves as a benchmark, helping the model better correct prediction errors. In subsequent training stages, by integrating physical depth with other self-supervised signals (such as photometric and geometric consistency), the model can further optimize depth prediction results. Ultimately, the foundational knowledge provided by physical depth significantly enhances the model's depth prediction accuracy and robustness, leading to excellent performance in practical applications.

% \begin{equation}
% \centering
% \begin{array}{l}
% {I_{t - 1 \to t}} = {I_{t - 1}}\left\langle {proj\left( {{D_t},{T_{t \to t - 1}}} \right)} \right\rangle \\
% co{n_{i,j}} = confidence\left( {{I_t},{I_{t - 1 \to t}}} \right)
% \end{array}
% \end{equation}
% where \(I_{t - 1 \to t}\) uses the image \(I_{t - 1}\)  to reconstruct the image at moment \(t\), and \(I_{t - 1}\) is the image at moment \(t-1\). The function \(\left\langle {proj\left( \cdot \right)} \right\rangle\) reconstructs the image at moment \(t\) from the image at moment \(t-1\) based on the depth and pose network. The function \(confidence\left( \cdot \right)\) calculates the confidence level for each physics depth based on \(I_t\) and the reconstructed \(I_{t - 1 \to t}\).

% In our study, the model prioritizes pixels with reliable depth data, enhancing overall depth accuracy. Conversely, it de-emphasizes low-confidence depths, reducing errors in high-confidence areas while tolerating imprecision in less reliable regions. To achieve this, we employ the L2 loss function, also known as Mean Squared Error (MSE). This function is essential for quantifying the discrepancy between the model's predicted depths and the actual depths across different coordinates. By integrating the confidence level into the loss calculation, we align the model's performance with the reliability of the depth data, prioritizing more accurate depth information.

\vspace{-3mm}
\begin{equation}
\centering
\begin{array}{l}
{L_{phy}} = \sum\nolimits_{i = 1}^M {\sum\nolimits_{j = 1}^N {{{\left( {d_{ij}^{phy} - {{\hat d}_{ij}}} \right)}^2}} }  
\end{array}
\vspace{-2mm}
\end{equation}
$d_{ij}^{phy}$ is the physics depth of $\left( {i,j} \right)$ as a label for supervision pixel point as a label for supervised learning. ${\hat d_{ij}}$ is the depth of $\left( {i,j} \right)$ predicted by the model.

Our model uses physics depth as its starting point, which helps in accurately predicting real depth. This is important because estimating depth from a single camera view often leads to scale issues, where the same point can appear to have different depths. Most single-view models can only estimate depth relative to other points, not the actual depth. However, our method uses physics depth during training, enabling the model to learn and correct errors in depth prediction. Since physics depth represents the true depth, it maintains scale accuracy during these corrections. Thus, our model effectively overcomes the scale problem in single depth estimation.

%-------------------------------------------------------------------------

\subsection{Self-Supervised Training}
In the self-supervised training paradigm, depth estimation is framed as an image reconstruction problem. This approach avoids the need for ground truth labels by utilizing unlabeled monocular videos during training. Our methodology leverages both photometric and geometric consistencies as dual pillars to jointly optimize image reconstruction.

% This dual-consistency approach allows for a more robust and accurate estimation of depth by capitalizing on the complementary strengths of both photometric and geometric information.

%-------------------------------------------------------------------------
\subsubsection{Photometric Consistency}
For consecutive frames \(I_{t-1}\) and \(I_t\), our model independently estimates their corresponding depths, \(D_{t-1}\) and \(D_t\). As outlined in Eq. \ref{project}, frames \(I_{t-1}\) and \(I_t\) can be projected into structured 3D point clouds \(Q_{t - 1}\) and \(Q_t\), respectively. Utilizing the pose network, we estimate the camera's motion from time \(t-1\) to \(t\). Through the application of the transformation matrix \(T_{t- 1 \to t}\) and the point cloud \(Q_t\), an estimated version of \(Q_{t-1}\), denoted as \({\hat Q}_{t-1}\), is obtained as \({\hat Q}_{t-1} = T_{t - 1 \to t} Q_t\). Subsequently, frame \(I_t\) is reconstructed by warping \(I_{t-1}\) using the principles detailed in Eq. \ref{photo}. The photometric loss is computed by Eq. \ref{photo} using reconstructed target image ${{I_{t- 1 \to t }}}$ and target image ${{I_t}}$. 
\vspace{-1.8mm}
\begin{equation}
\centering
\begin{array}{l}
Q_{t-1}^{xy} = D_{t-1}^{xy} \cdot {K^{ - 1}}\left[ {\begin{array}{*{20}{c}} \label{project}
x\\
y\\
1
\end{array}} \right]
\end{array}
\end{equation}
\vspace{-4mm}
\begin{equation}
\centering
\begin{array}{l}
{I_{t -1\to t}}\left[ u \right] = {I_{t-1}\left\langle {u'} \right\rangle } ,
{L_{ph}} = ph\left( {{I_t},{I_{t - 1 \to t}}} \right) \label{photo}
\vspace{-3mm}
\end{array}
\end{equation}
\vspace{-3mm}
\begin{equation}
\begin{array}{l}
ph\left( {{I_t},{I_{t - 1 \to t}}} \right) = \frac{\alpha }{2}\left( {1 - SSIM\left( {{I_t},{I_{t - 1 \to t}}} \right)} \right)  \\
+\left( {1 - \alpha } \right){\left\| {\left( {{I_t},{I_{t - 1 \to t}}} \right)} \right\|_1}
\end{array}
\end{equation}
Here $\alpha$ is commonly set to 0.85 \cite{godard2019digging}, ph is a photometric reconstruction error. Furthermore, for each pixel p, the minimum of the losses computed from forward and backward neighboring frames allows the mitigation of the effect of occlusions \cite{godard2019digging} on the reprojection process.
\begin{equation}
\centering
\begin{array}{l}
{L_{ph}}\left( p \right) =\mathop {\min }\limits_{s \in \left[ { - 1,1} \right]} pe\left( {{I_{t-1}}\left( p \right),{I_{{t-1} \to t}}\left( p \right)} \right)
\end{array}
\end{equation}
$1$ stands for forward, $-1$ stands for backward.
% \begin{equation}
% \centering
% \begin{array}{l}
% {L_s} = \left| {{\partial _x}d_t^*} \right|{e^{ - \left| {{\partial _x}{I_t}} \right|}} + \left| {{\partial _y}d_t^*} \right|{e^{ - \left| {{\partial _y}{I_t}} \right|}}
% \end{array}
% \end{equation}
% As in previous works \cite{godard2019digging}, the edge-aware smoothness loss is used to improve the depth through sharpening the edges and smoothing the continuous depth surfaces.
%-------------------------------------------------------------------------

\begin{table*}[t]
\begin{center}
\resizebox{\textwidth}{!}{
\begin{tabular}{llllllllll}
\toprule
\multicolumn{1}{c}{\bf Method}  &\multicolumn{1}{c}{\bf Scale}  &\multicolumn{1}{c}{\bf Test } &\multicolumn{1}{c}{\bf AbsRel ↓} &\multicolumn{1}{c}{\bf Sq Rel↓}   &\multicolumn{1}{c}{\bf RMSE↓} &\multicolumn{1}{c}{\bf RMSElog ↓} &\multicolumn{1}{c}{\bf $\delta < 1.25$\ ↑}  &\multicolumn{1}{c}{\bf $\delta < 1.25^2$\ ↑}  &\multicolumn{1}{c}{\bf $\delta < 1.25^3$\ ↑} 
\\ 
\midrule

Monodepth2 \cite{godard2019digging} &LiDAR Scale & 32.260 & 0.159 & 1.689 & 5.168 & 0.238 & 0.830 & 0.931 & 0.967 
\\  
 &Physics Depth Scale & 32.487 & 0.158 & 1.968 & 5.287 & 0.242 & 0.842 & 0.930 & 0.966 
\\   
MonoVit \cite{zhao2022monovit}&LiDAR Scale & 28.354 & 0.110 & 0.759 & 4.248 & 0.199 & 0.872 & 0.954 & 0.979 
\\  
&Physics Depth Scale & 28.096 & 0.108 & 0.743 & 4.241 & 0.200 & 0.874 & 0.955 & 0.979 

\\    
SQLDepth \cite{wang2023sqldepth}        &LiDAR Scale&43.51 &0.087  &0.659 &4.096 &0.165 &0.920 &0.970 &0.984
\\    
    &Physics Depth Scale&44.17 &0.089  &0.664 &4.101 &0.169 &0.918 &0.969 &0.982
\\   
\bottomrule
\end{tabular}}
\end{center}
\vspace{-3mm}
\caption{ Evaluation of different models with LiDAR Depth Scaling Factor and Physics Depth Scaling Factor. } 
\vspace{-6mm}
\label{scale}
\end{table*}

\begin{table*}[h]
\begin{center}
\resizebox{\textwidth}{!}{
\begin{tabular}{lllll}
\toprule
KITTI Date  & \begin{tabular}[c]{@{}l@{}}Road Physics Depth\\Error: +/- 5\%\end{tabular} & \begin{tabular}[c]{@{}l@{}}Road Physics Depth\\Error: +/- 10\%\end{tabular} & \begin{tabular}[c]{@{}l@{}}Flat Surface Physics Depth\\Error: +/- 5\%\end{tabular} & \begin{tabular}[c]{@{}l@{}}Flat Surface Physics Depth\\Error: +/- 10\%\end{tabular} \\
\midrule
2011-09-26 &84.28\% &96.26\% &75.08\% &89\%\\
2011-09-28 &80.61\% &85.64\% &61.21\% &77\%\\
2011-09-29 &90.53\% &97.34\% &74.46\% &91\%\\
2011-09-30 &76.43\% &91.86\% &56.98\% &81\%\\
2011-10-0  &78.12\% &94.61\% &62.77\% &85\%\\

\bottomrule
\end{tabular}}
\end{center}
\vspace{-3mm}
\caption{\textbf{Error between physics depth and KITTI ground truth.} The proportion of the 5-days road physics depth error and the flat surface physics depth error within $5\%$ and within $10\%$ of ground truth, respectively, in the KITTI dataset. }
\vspace{-6mm}
\label{kitti error}
\end{table*}

\begin{table}[t]
\begin{center}
\resizebox{0.47\textwidth}{!}{
\begin{tabular}{lllll}
\toprule
City & \begin{tabular}[c]{@{}l@{}}Road Physics Depth\\Error: +/- 5\%\end{tabular} & \begin{tabular}[c]{@{}l@{}}Road Physics Depth\\Error: +/- 10\%\end{tabular} & \begin{tabular}[c]{@{}l@{}}Flat Surface Physics Depth\\Error: +/- 5\%\end{tabular} & \begin{tabular}[c]{@{}l@{}}Flat Surface Physics Depth\\Error: +/- 10\%\end{tabular} \\
\midrule
aachen & 87.48\% & 94.77\% & 73.17\% & 86.94\% \\
bochum & 80.76\% & 93.22\% & 65.51\% & 83.95\% \\
bremen & 86.55\% & 97.64\% & 72.60\% & 88.29\% \\
cologne & 81.66\% & 98.88\% & 75.14\% & 88.82\% \\
darmstadt & 82.49\% & 95.44\% & 69.95\% & 86.56\% \\
dusseldorf & 83.22\% & 93.59\% & 68.79\% & 84.96\% \\
erfurt & 83.78\% & 94.26\% & 69.58\% & 85.85\% \\
hamburg & 82.77\% & 96.81\% & 67.93\% & 84.22\% \\
hanover & 76.59\% & 97.45\% & 64.71\% & 83.00\% \\
monchengladbach & 83.42\% & 94.73\% & 63.75\% & 82.48\% \\
strasbourg & 84.63\% & 95.62\% & 61.44\% & 81.52\% \\
stuttgart & 80.49\% & 96.38\% & 68.52\% & 85.26\% \\
tubingen & 85.44\% & 92.76\% & 67.22\% & 84.69\% \\
ulm & 89.00\% & 98.38\% & 73.35\% & 87.89\% \\
weimar & 80.06\% & 93.69\% & 64.47\% & 82.58\% \\
zurich & 88.99\% & 97.52\% & 70.72\% & 85.82\% \\
jena & 77.90\% & 92.85\% & 63.75\% & 81.85\% \\
krefeld & 86.23\% & 94.11\% & 65.83\% & 83.92\% \\
\bottomrule
\end{tabular}}
\end{center}
\vspace{-3mm}
\caption{\textbf{Error between physics depth and Cityscape ground truth:} The proportion of Road Physics Depth Error and Flat Surface Physics Depth Error for different cities in the cisyscape dataset. }
\vspace{-5mm}
\label{cityscape error}
\end{table}

\begin{table}[t]
\resizebox{0.49\textwidth}{!}{
\begin{tabular}{lll}
\toprule
               & Road Physics Depth & Ground PhysicsDepth \\ 
\midrule
+/- 5\% error  & 80.24\%                & 60.30\%    \\ 
+/- 10\% error & 99.33\%                & 74.89\%   \\ 
\bottomrule
\end{tabular}}
\caption{Physics depth in a sample KITTI image. }
\label{image error}
\vspace{-7mm}
\end{table}

\begin{table*}[t]
\begin{center}
\resizebox{\textwidth}{!}{
\begin{tabular}{lllllllllll}
\toprule
\multicolumn{1}{c}{\bf Method}   &\multicolumn{1}{c}{\bf Type} &\multicolumn{1}{c}{\bf Year} &\multicolumn{1}{c}{\bf Resolution}  &\multicolumn{1}{c}{\bf AbsRel ↓} &\multicolumn{1}{c}{\bf Sq Rel↓}   &\multicolumn{1}{c}{\bf RMSE↓} &\multicolumn{1}{c}{\bf RMSE log↓} &\multicolumn{1}{c}{\bf $\delta < 1.25$\ ↑}  &\multicolumn{1}{c}{\bf $\delta < 1.25^2$\ ↑}  &\multicolumn{1}{c}{\bf $\delta < 1.25^3$\ ↑} \\ 
\midrule
Monodepth2 \cite{godard2019digging}   &MS&2019 &1024×320 &0.106 &0.806 &4.630 &0.193 &0.876 &0.958 &0.980\\  
HR-Depth \cite{lyu2021hr}            &MS&2021 &1024×320 &0.101 &0.716 &4.395 &0.179 &0.899 &0.966 &0.983\\  
Lite-Mono \cite{zhang2023lite}         &M &2023&1024×320 &0.097 &0.710 &4.309 &0.174 &0.905 &0.967 &0.984 \\  
MonoVIT \cite{zhao2022monovit}        &M &2023 &1024×320 &0.096 &0.714 &4.292 &0.172 &0.908 &0.968 &0.984\\  
DualRefine \cite{bangunharcana2023dualrefine}    &MS &2023&1024×320 &0.096 &0.694 &4.264 &0.173 &0.908 &0.968 &0.984\\  
ManyDepth \cite{watson2021temporal}  &M &2021 &1024×320&0.087 &0.685 &4.142 &0.167 &0.920 &0.968 &0.983\\  
RA-Depth \cite{he2022ra}  &M &2022&1024×320 &0.097 &0.608& 4.131& 0.174 &0.901 &0.968& 0.985\\  
PlaneDepth \cite{wang2023planedepth}  &MS &2023 &1280×384  &0.090 &0.584 &4.130 &0.182 &0.896& 0.962 &0.981\\  
SQLDepth   \cite{wang2023sqldepth}        &M &2023&1024×320 &0.087  &0.659 &4.096 &0.165 &0.920 &0.970 &0.984\\  
% \\  \hline 
% SfM-TTR\cite{izquierdo2023sfm} &M &2023 &1024×320&0.087 &0.660 &3.948 &0.165 &0.925 &0.969 &0.984
Ours (Backbone: SQLDepth)        &M &2023 &1024×320 &   0.085  &   0.583  &   3.885  &   0.158  &   0.922  &   0.970  &   0.986\\ 
\bottomrule
\end{tabular}}
\end{center}
\vspace{-3mm}
\caption{The quantitative depth comparison using the Eigen split of the KITTI dataset ~\protect\cite{geiger2013vision}. M: trained with monocular videos; MS: trained with stereo pairs.}
\vspace{-6mm}
\label{kitty_result}
\end{table*}

\begin{table*}[t]
\begin{center}
\resizebox{\textwidth}{!}{
\begin{tabular}{llllllllll}
\toprule
\multicolumn{1}{c}{\bf Method}  &\multicolumn{1}{c}{\bf Size}  &\multicolumn{1}{c}{\bf Test } &\multicolumn{1}{c}{\bf AbsRel ↓} &\multicolumn{1}{c}{\bf Sq Rel↓}   &\multicolumn{1}{c}{\bf RMSE↓} &\multicolumn{1}{c}{\bf RMSElog ↓} &\multicolumn{1}{c}{\bf $\delta < 1.25$\ ↑}  &\multicolumn{1}{c}{\bf $\delta < 1.25^2$\ ↑}  &\multicolumn{1}{c}{\bf $\delta < 1.25^3$\ ↑} \\
\midrule
Pilzer et al \cite{pilzer2018unsupervised}      &$512\times\ 256$ &1 &0.240 &4.264 &8.049& 0.334 &0.710 &0.871 &0.937\\  
Struct2Depth \cite{casser2019unsupervised}    &$416\times\ 128$ &1 &0.145 &1.737 &7.280 &0.205  &0.813 & 0.942 &0.976\\    
Monodepth2 \cite{godard2019digging} &$416\times\ 128$ &1 &0.129 &1.569 &6.876 &0.187  & 0.849 &0.957 &0.983\\  
Lee \cite{lee2021attentive} &$832\times\ 256$ &1 &0.111 &1.158 &6.437 &0.182 &0.868 &0.961 &0.983\\  
InstaDM \cite{lee2021learning} &$832\times\ 256$  &1 &0.111 &1.158 &6.437 &0.182 &0.868 &0.961 &0.983\\   
ManyDepth \cite{watson2021temporal}   &$416\times\ 128$ &2  &0.114 &1.193 &6.223 &0.170 &0.875 &0.967 & 0.989\\   
SQLDepth \cite{wang2023sqldepth}   &$416\times\ 128$ &1 &0.110 &1.130 &6.264 &0.165 &0.881 &0.971 &0.991\\    
Ours (Backbone: SQLDepth) &$416\times\ 128$  &1 &0.103 &1.090 &6.237 &0.157 &0.895 &0.974 &0.991\\   
\bottomrule
\end{tabular}}
\end{center}
\vspace{-3mm}
\caption{The quantitative depth comparison of the Cityscape dataset. } 
\vspace{-9mm}
\label{city_result}
\end{table*}

\subsubsection{2D Spatial Consistency}

Our method prioritizes spatial disparities to gauge scene geometry, which is key for decoding object interactions. This strategy excels in depth and scale depiction, functioning well in scenes with sparse texture or color and remaining stable against lighting or color shifts. We devised a model that refines pose and depth networks via a loss function, using spatial disparities to boost pose accuracy and depth perception. By utilizing dense optical flow, the model aligns points across frames to compute their movement, aiding in precise motion analysis. Our loss function, ${L_{2D}}$, is based on motion variance between actual and reconstructed frames, ${I_t}$ and ${\hat I_t}$, enhancing model accuracy.

\vspace{-4mm}
\begin{equation}
{L_{2D}} = \sum_{i=1}^{N} \left( \alpha \|\mathbf{v}_{t+1}^{(i)} - \mathbf{v}_{t}^{(i)}\|^2 + \beta (1 - \cos(\theta^{(i)})) \right)
\end{equation}

$N$ represents the total number of matching points. $alpha$ and $beta$ are weight coefficients balancing the positional and directional differences. ${v}_{t}^{(i)}$ and $\mathbf{v}_{t+1}^{(i)}$ denote the motion vectors of the $i-th$ matching point in frames $t $ and $ t+1 $, respectively. \( \theta^{(i)} \) is the angle between vectors \( \mathbf{v}_{t}^{(i)} \) and \( \mathbf{v}_{t+1}^{(i)} \), with \( \cos(\theta^{(i)}) = \frac{\mathbf{v}_{t}^{(i)} \cdot \mathbf{v}_{t+1}^{(i)}}{\|\mathbf{v}_{t}^{(i)}\| \|\mathbf{v}_{t+1}^{(i)}\|} \) representing the cosine of this angle. The first term in the loss function, \( \|\mathbf{v}_{t+1}^{(i)} - \mathbf{v}_{t}^{(i)}\|^2 \), quantifies the positional difference, whereas the second term, \( 1 - \cos(\theta^{(i)}) \), assesses the directional difference. The directional consistency is measured by the cosine of the angle between the motion vectors, where a value close to 1 indicates minimal directional change, and a value further from 1 signifies a greater change. 
% The parameters \( \alpha \) and \( \beta \) allow for fine-tuning the relative importance of positional versus directional discrepancies in the overall loss computation.

\begin{table}[h]
\begin{center}
\resizebox{0.49\textwidth}{!}{
\begin{tabular}{llllllllll}
\toprule
\multicolumn{1}{c}{\bf Method}  &\multicolumn{1}{c}{\bf Type} &\multicolumn{1}{c}{\bf AbsRel ↓} &\multicolumn{1}{c}{\bf Sq Rel↓}   &\multicolumn{1}{c}{\bf RMSE↓} &\multicolumn{1}{c}{\bf log10↓} \\  
\midrule
Zhou \cite{zhou2017unsupervised}  &S &0.383 &5.321 &10.470 &0.478 \\ 
DDVO \cite{wang2018learning}   &M &0.387 &4.720 &8.090 &0.204 \\ 
Monodepth2 \cite{godard2019digging}    &M &0.322 &3.589 &7.417 &0.163\\  
CADepthNet \cite{yan2021channel}    &M &0.312 &3.086 &7.066 &0.159\\         
SQLDepth \cite{wang2023sqldepth}   &M &0.306 &2.402& 6.856 &0.151\\    
Ours (Backbone: SQLDepth)   &M &0.304 &2.313 &6.822 &0.148\\   
\bottomrule
\end{tabular}}
\end{center}
\vspace{-3mm}
\caption{ The quantitative depth comparison of the Make3d. } 
\vspace{-10mm}
\label{Make3d_result}
\end{table}

\begin{table*}[t]
\begin{center}
\resizebox{1\textwidth}{!}{
\begin{tabular}{lllllllll}
\toprule
\multicolumn{1}{c}{\bf {Back-bone }}  &\multicolumn{1}{c}{\bf \parbox{1cm}{Physics \\ Depth}}  &\multicolumn{1}{c}{\bf confidence}  &\multicolumn{1}{c}{\bf \parbox{2cm}{2D Spatial\\ Consistency}}  &\multicolumn{1}{c}{\bf AbsRel ↓} &\multicolumn{1}{c}{\bf Sq Rel↓}   &\multicolumn{1}{c}{\bf RMSE↓} &\multicolumn{1}{c}{\bf RMSE log↓} &\multicolumn{1}{c}{\bf $\delta < 1.25$\ ↑} \\  
\midrule
\checkmark &   & &   &0.087 &0.659 &4.096 &0.165 &0.920\\  
\checkmark &\checkmark    &  & &0.086  &   0.621  &   3.912  &   0.161  &   0.921\\  
\checkmark &\checkmark &\checkmark &   &0.086 &0.594 &3.886 &0.159 &0.918\\  
 \checkmark &  & & \checkmark &     0.087  &   0.620  &   4.043  &   0.164  &   0.913   \\    
\checkmark &\checkmark  &\checkmark & \checkmark &  0.085  &   0.583  &   3.885  &   0.158  &   0.922  \\ 
\bottomrule
\end{tabular}}
\end{center}
\vspace{-2mm}
\caption{Ablation study on KITTI: Input is $1024\times320$. ${L_{con}}$: Loss of Physics-Depth Supervision. ${L_{2D}}$: Loss of 2D Spatial Consistency. }
\vspace{-6mm}
\label{Ablation_result}
\end{table*}

\section{Experiment}

\subsection{Implementation Details}
The depth estimation network framework references SQLdepth \cite{wang2023sqldepth}. The pose network, PoseCNN, receives 3 frames as input and outputs axis-angle and translation components to describe the change in camera pose. The network consists of 7 convolutional layers, each followed by a ReLU activation function. 
The output of the convolutional layers goes through a 1x1 convolutional layer and then undergoes an average pooling operation to obtain the pose estimation vector.
The model is trained on 4 NVIDIA A6000 GPUs.
Our training process is divided into two phases: a supervised learning phase and a self-supervised learning phase. During the supervised learning phase, we use physical depth as labels and train the model for 15 epochs. In the subsequent self-supervised learning phase, the model is trained for 20 epochs.
We use the Adam optimizer to jointly train DepthNet and PoseNet with parameters $\beta_1=0.9$ and $\beta_2=0.999$. The initial learning rate is set to $1 \times 10^{-4}$ and decays to $1 \times 10^{-5}$ after 15 epochs. We set the SSIM weight to $\alpha=0.85$ and the smooth loss term weight to $\lambda=1 \times 10^{-3}$. 
% We use ResNet-$50$ \cite{he2016deep} with ImageNet \cite{russakovsky2015imagenet} pretrained weights as the backbone network.

\subsection {Physics Depth Evaluation}
\label{section:Physics Depth Demonstration}
\begin{figure}[h]
\begin{center}
%\framebox[4.0in]{$\;$}
\includegraphics[width=8.5cm, height=5cm]{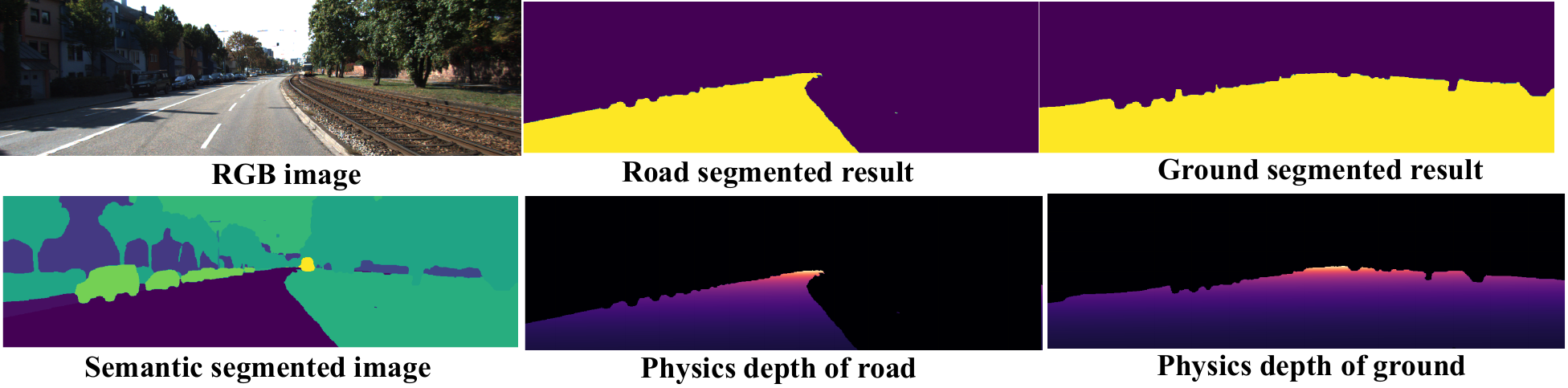}
\end{center}
\vspace{-6mm}
\caption{ Physics Depth Methodology demonstrated on KITTI. }
\vspace{-4mm}
\label{Depth Methodology}
\end{figure}

\textbf{Physics Depth Methodology:} 
In Section \ref{sec:formatting}, we explore two different types of physics depth: road physics depth and ground physics depth. Utilizing the KITTI dataset, we visualize the outcomes of these different types of physics depth in Figure \ref{Depth Methodology}. 'd' shows the semantic segmentation results obtained using a pre-trained segmentation model, while 'e' and 'f' represent the visualized results of the physics depth calculated for the road and ground areas.

\begin{figure}[t]
\begin{center}
%\framebox[4.0in]{$\;$}
\includegraphics[width=8.5cm, height=4cm]{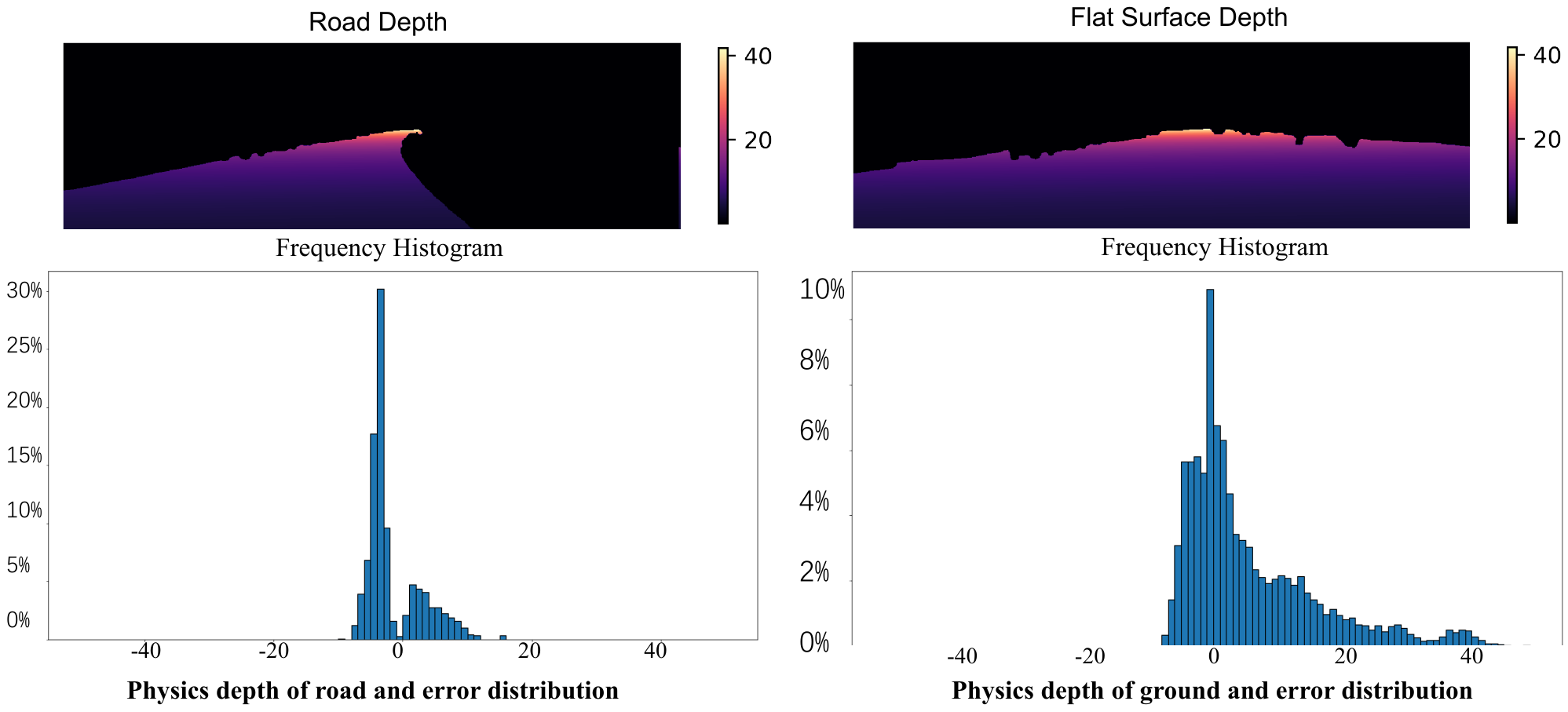}
\vspace{-6mm}
\end{center}
\caption{Error distribution of Physics depth. }
\vspace{-6mm}
\label{Physics Depth Ablation}
\end{figure}

\textbf{Error Distribution:} The comparison of physics depth logic on a sample image, shown in Fig. \ref{Physics Depth Ablation} and Table \ref{image error}, highlights its effectiveness in estimating road surfaces. With over 99\% of pixels showing less than 10\% error and more than 81\% exhibiting less than 5\% error compared to LiDAR, it confirms the potential of physics depth estimations for flat surfaces like roads as a viable LiDAR alternative for scaling factor calculations in self-supervised monocular depth estimation algorithms. However, accuracy declines with non-level surfaces like sidewalks or rail tracks due to their variable elevation relative to the camera's base. Extending the logic to vertical surfaces slightly increases error, a trade-off for denser physics depth estimation.

\textbf{Scale Alignment:} In Table \ref{scale}, we compared three monocular depth estimation models by calculating ratios of model-predicted depths to both ground truth and physics depth. After adjusting the predicted depths with these ratios, we assessed the models against ground truth metrics. Results show that scales from physics depth closely match those from ground truth, with performance metrics nearly identical. Notably, the MonoVit model sometimes surpassed the performance of scales based on ground truth. This confirms that physics depth are reliable alternatives to LiDAR depths for scaling factor calculations, significantly improving the autonomy of self-supervised models.

% \begin{figure}[t]
% \begin{center}
% \includegraphics[width=8cm, height=4cm]{image/percentage_error.png}
% \end{center}
% \vspace{-8mm}
% \caption{Percentage error frequency distribution for KITTI dataset. The top row of the figure showcases the accuracy of the physics depth, presenting a frequency distribution of percentage errors for road pixels across the complete KITTI dataset. This distribution enables a direct comparison between the physics depth and KITTI's LiDAR depth for the specified timestamps: (a) 2011-09-26, (b) 2011-09-28, (c) 2011-09-29, (d) 2011-09-30, and (e) 2011-10-03. The second row illustrates the analogous error distributions after corrections have been applied to KITTI’s camera calibration.}
% \vspace{-4mm}
% \label{percentage}
% \end{figure}

\begin{figure*}[h]
%\vspace{-0.4cm}
\begin{center}
\includegraphics[width=17.4cm, height=5cm]{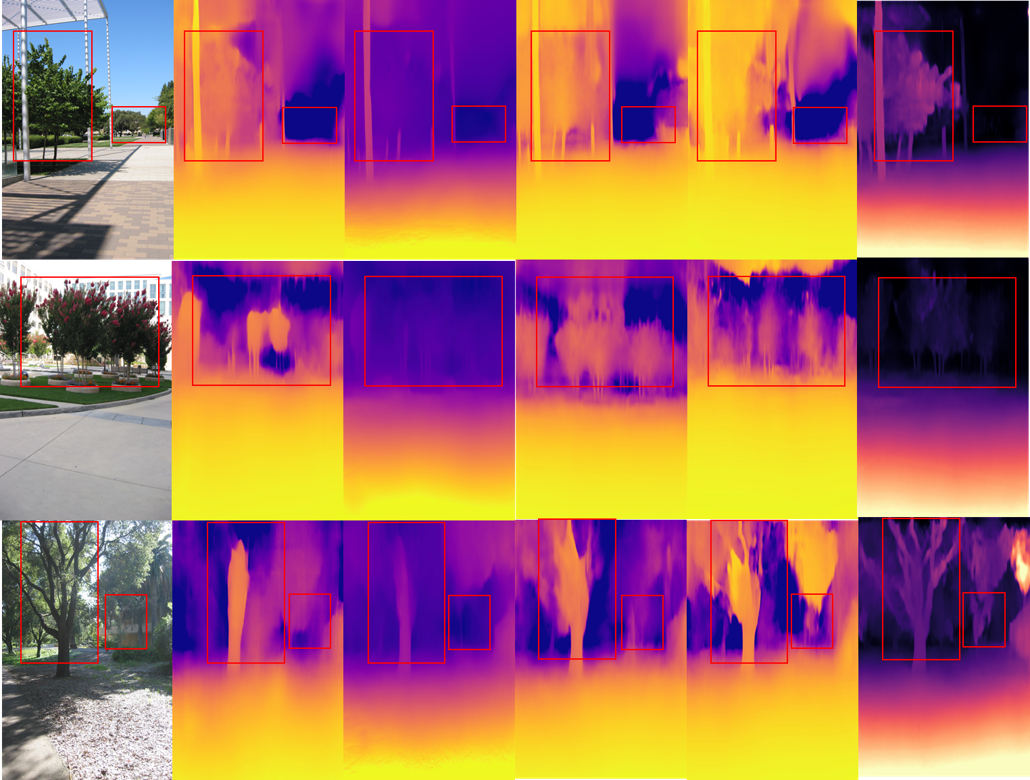}
\end{center}
\vspace{-5mm}
\caption{Qualitative results on make3d (Zero-shot): From left to right the models are Monodepth2 ~\protect\cite{godard2019digging}, RA-Depth ~\protect\cite{he2022ra}, MonoVit~\protect\cite{zhao2022monovit},  SQLDepth~\protect\cite{wang2023sqldepth}, our models.}
\vspace{-4mm}
\label{qualitative of make3d}
\end{figure*}

\begin{figure}[h]
%\vspace{-0.4cm}
\begin{center}
\includegraphics[width=8.5cm, height=5cm]{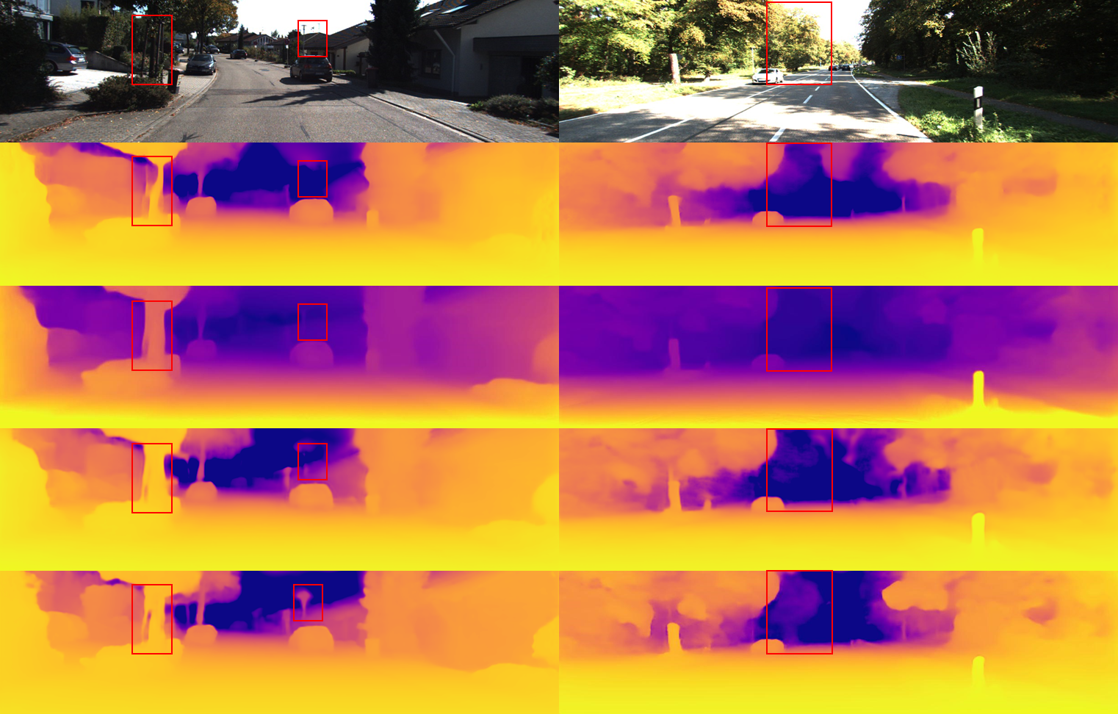}
\end{center}
\vspace{-5mm}
\caption{ Qualitative results on KITTI: From top to bottom the models are MonoVit~\protect\cite{zhao2022monovit}, RA-Depth ~\protect\cite{he2022ra}, ManyDepth~\protect\cite{watson2021temporal}, our models.}
\vspace{-5mm}
\label{kitti2}
\end{figure}

\vspace{-2mm}
\subsection {Evaluation of Physics Depth}
\vspace{-1mm}
In this paper, we have systematically generated physics depth for the entire KITTI and Cityscapes datasets to facilitate the training of our models. This involved a meticulous analysis of the discrepancies in both road and flat surface physics depth across these datasets. As detailed in Tables \ref{kitti error} and \ref{cityscape error}, the KITTI dataset showed approximately \(90\%\) of pixels exhibited an error margin of less than \(10\%\), and about \(80\%\) of pixels were within a mere \(5\%\) deviation when compared with the LiDAR-generated depth. Notably, the Cityscapes dataset demonstrated exceptional performance. In this dataset, around \(95\%\) of pixels showed less than a \(10\%\) error margin, and \(85\%\) of pixels were within a \(5\%\) error range, in comparison to the depth derived from Cityscapes' standard disparity data.

Tables \ref{kitti error} and \ref{cityscape error} indicate that the road physics depth outperforms the flat surface physics depth in accuracy. Yet, road pixels in a single image are limited. To increase the density of physics depth pixels in each image, we applied the logic to flat surfaces, although the flatness of these surfaces is not exactly the same. This extension, while increasing data, also enlarges the error margin with the ground truth. Still, as seen in Tables \ref{kitti error} and \ref{cityscape error}, the flat surface physics depth, despite higher errors, maintains good accuracy, enriching the dataset and reducing the risk of overfitting.

Our analysis showed that the KITTI dataset had lower accuracy than Cityscapes, likely due to differences in camera calibration quality. KITTI uses one calibration file per day, while Cityscapes has individual files for each image, suggesting that better calibration enhances physics depth accuracy. This implies that improved calibration could further increase the accuracy of physics depth. Our physics depth estimation method, especially for flat surfaces like roads, shows promising potential, replacing LiDAR in calculating scale factors for self-supervised monocular depth estimation. 
% \begin{figure*}[t]
% %\vspace{-0.4cm}
% \begin{center}
% \includegraphics[width=14cm, height=9cm]{image/kitti_result.jpg}
% \end{center}
% \caption{ Qualitative results on KITTI: From top to bottom the models are MonoVit~\protect\cite{zhao2022monovit}, RA-Depth ~\protect\cite{he2022ra}, ManyDepth~\protect\cite{watson2021temporal}, our models.}

% \label{qualitative}
% \end{figure*}

\vspace{-3mm}
\subsection{Depth Estimation}

\textbf{KITTI:} 
Our model was assessed on the KITTI Eigen split of 697 images, and the results, displayed in Table \ref{kitty_result}, demonstrate that our method significantly surpasses existing self-supervised techniques on the KITTI dataset. Our advancements, such as physics depth, confidence metrics, and 2D consistency checks, have notably enhanced performance, particularly in RMSE metrics. Figure \ref{kitti2} illustrates our model's exceptional capability in capturing complex scene details and reconstructing scenes more accurately than other models like MonoVit, RA-Depth, and ManyDepth.

\textbf{Cityscapes:} 
To assess the generalizability of our model, we showcase results from the Cityscapes dataset. In Table \ref{city_result}, we perform additional comparisons where we train and test on the Cityscapes dataset. We consistently outperform competing methods.

\textbf{Make3D:} 
We evaluated our model's generalization capabilities through a zero-shot test on the Make3D dataset, using a version pretrained on KITTI. Results in Table \ref{Make3d_result} show that our model achieves lower errors than other zero-shot competitors, highlighting its superior zero-shot generalization. Figure \ref{qualitative of make3d} demonstrates that our model surpasses baseline models, producing high-quality depth with improved sharpness and scene detail accuracy, proving its exceptional ability to adapt to new scenarios without further fine-tuning.

% \textbf{3D reconstruction: }  This paper presents a model that predicts depth from single images for 3D reconstruction, as demonstrated in Fig. \ref{house}. Our model's key strength is its accurate depth prediction, enhancing overall reconstruction quality, particularly in detailed areas like windows, which are reconstructed as regular rectangles compared to the distorted versions by other models. It also excels in depicting distant structures like houses with minimal distortion, underscoring its superior depth prediction and 3D reconstruction capabilities compared to competitors.

\vspace{-2mm}
\subsection {Ablation Study}
\vspace{-1.5mm}
Our ablation study, presented in Table \ref{Ablation_result}, evaluates the impact of various components in monocular depth estimation. Results indicate that integrating all components enhances performance compared to the baseline model.

\textbf{Physics Depth:} Table \ref{Ablation_result} shows that accurate ground depth significantly improves depth prediction, which is essential for precise object positioning and overall scene understanding. This enhancement in spatial perception not only clarifies size and distance ambiguities but also accelerates the model's training and its ability to adapt to the geometry of diverse scenes.

\textbf{Confidence in Physics Depth:} As shown in Table \ref{Ablation_result}, our method, which assigns confidence scores to physics depth estimation, surpasses basic physics depth. Incorporating confidence scores enables our model to focus on more accurate regions during training, reducing error impact and improving self-supervision efficacy.

\textbf{2D Spatial Consistency:} Table \ref{Ablation_result} indicates that our model, which utilizes optical flow for 2D reprojection error calculation, outperforms the baseline model that relies solely on photometric consistency.

Our ablation study reveals that incorporating Physics Depth significantly enhances monocular depth estimation accuracy by initiating training with accurate ground depth. This self-supervised method, which emphasizes reliable Physics Depth areas, outperforms traditional approaches. Additionally, adding 2D Spatial Consistency further boosts accuracy.

% \begin{figure*}[t]
% %\vspace{-0.4cm}
% \begin{center}
% \includegraphics[width=16cm, height=4cm]{image/make3d.png}
% \end{center}
% \vspace{-5mm}
% \caption{Qualitative results on make3d (Zero-shot): From left to right the models are Monodepth2 ~\protect\cite{godard2019digging}, RA-Depth ~\protect\cite{he2022ra}, MonoVit~\protect\cite{zhao2022monovit},  SQLDepth~\protect\cite{wang2023sqldepth}, our models.}
% \vspace{-3mm}
% \label{qualitative of make3d}
% \end{figure*}
% \vspace{-2mm}
\vspace{-1mm}
\section{Conclusion}
% This paper presents a novel self-supervised learning approach that improves monocular depth estimation using physics-based depth cues. Our method surpasses existing self-supervised techniques by incorporating physics depth estimation, enhancing accuracy and environmental detail capture. It allows models to accurately predict scene depth and 3D structures, significantly advancing self-supervised learning with state-of-the-art results on KITTI, Cityscapes, and Make3D datasets. This approach yields better depth information for improved understanding of real-world scenes.
This paper presents a novel self-supervised learning approach that integrates the physical characteristics of the camera model with the concept of embodiment, embedding them into the deep learning model. By leveraging physics-based depth cues, our method improves monocular depth estimation. Through the incorporation of physics depth estimation, our approach achieves an embodied understanding of the interaction between the camera and the physical world, enhancing the model's accuracy and its ability to capture environmental details, surpassing existing self-supervised techniques. This method enables the model to accurately predict scene depth, achieving state-of-the-art self-supervised learning results on the KITTI, Cityscapes, and Make3D datasets. This approach provides better depth information, thereby deepening the model's embodied understanding of real-world scenes.

 \section*{Acknowledgement} This work is supported by Ford Motor Company and NSF Awards No. 2334624, 2340882 and No. 2334690.

% \begin{figure}[t]
% %\vspace{-0.4cm}
% \begin{center}
% \includegraphics[width=8cm, height=3cm]{image/house3d.png}
% \end{center}
% \vspace{-5mm}
% \caption{ Qualitative results on KITTI 3D reconstruction: From left to right the models are ours, RA-Depth ~\protect\cite{he2022ra}, MonoVit~\protect\cite{zhao2022monovit},  Monodepth2~\protect\cite{godard2019digging}.}
% \vspace{-6mm}
% \label{house}
% \end{figure}

\vspace{-2mm}
{\small
\bibliographystyle{ieee_fullname}
\bibliography{egbib}

\begin{thebibliography}{10}\itemsep=-1pt

\bibitem{bangunharcana2023dualrefine}
Antyanta Bangunharcana, Ahmed Magd, and Kyung-Soo Kim.
\newblock Dualrefine: Self-supervised depth and pose estimation through iterative epipolar sampling and refinement toward equilibrium.
\newblock In {\em CVPR}, 2023.

\bibitem{bhat2021adabins}
Shariq~Farooq Bhat, Ibraheem Alhashim, and Peter Wonka.
\newblock Adabins: Depth estimation using adaptive bins.
\newblock In {\em CVPR}, 2021.

\bibitem{casser2019unsupervised}
Vincent Casser, Soeren Pirk, Reza Mahjourian, and Anelia Angelova.
\newblock Unsupervised monocular depth and ego-motion learning with structure and semantics.
\newblock In {\em CVPR Workshops}, 2019.

\bibitem{chawla2021multimodal}
Hemang Chawla, Arnav Varma, Elahe Arani, and Bahram Zonooz.
\newblock Multimodal scale consistency and awareness for monocular self-supervised depth estimation.
\newblock In {\em ICRA}. IEEE, 2021.

\bibitem{cordts2016cityscapes}
Marius Cordts, Mohamed Omran, Sebastian Ramos, Timo Rehfeld, Markus Enzweiler, Rodrigo Benenson, Uwe Franke, Stefan Roth, and Bernt Schiele.
\newblock The cityscapes dataset for semantic urban scene understanding.
\newblock In {\em CVPR}, 2016.

\bibitem{eigen2014depth}
David Eigen, Christian Puhrsch, and Rob Fergus.
\newblock Depth map prediction from a single image using a multi-scale deep network.
\newblock {\em NeurIPS}, 27, 2014.

\bibitem{gallup2010piecewise}
David Gallup, Jan-Michael Frahm, and Marc Pollefeys.
\newblock Piecewise planar and non-planar stereo for urban scene reconstruction.
\newblock In {\em CVPR}. IEEE, 2010.

\bibitem{garg2020wasserstein}
Divyansh Garg, Yan Wang, Bharath Hariharan, Mark Campbell, Kilian~Q Weinberger, and Wei-Lun Chao.
\newblock Wasserstein distances for stereo disparity estimation.
\newblock {\em NeurIPS}, 33, 2020.

\bibitem{garg2016unsupervised}
Ravi Garg, Vijay~Kumar Bg, Gustavo Carneiro, and Ian Reid.
\newblock Unsupervised cnn for single view depth estimation: Geometry to the rescue.
\newblock In {\em Computer Vision--ECCV 2016: 14th European Conference, Amsterdam, The Netherlands, October 11-14, 2016, Proceedings, Part VIII 14}. Springer, 2016.

\bibitem{geiger2013vision}
Andreas Geiger, Philip Lenz, Christoph Stiller, and Raquel Urtasun.
\newblock Vision meets robotics: The kitti dataset.
\newblock {\em IJRR}, 32, 2013.

\bibitem{godard2017unsupervised}
Cl{\'e}ment Godard, Oisin Mac~Aodha, and Gabriel~J Brostow.
\newblock Unsupervised monocular depth estimation with left-right consistency.
\newblock In {\em CVPR}, 2017.

\bibitem{godard2019digging}
Cl{\'e}ment Godard, Oisin Mac~Aodha, Michael Firman, and Gabriel~J Brostow.
\newblock Digging into self-supervised monocular depth estimation.
\newblock In {\em ICCV}, 2019.

\bibitem{guizilini20203d}
Vitor Guizilini, Rares Ambrus, Sudeep Pillai, Allan Raventos, and Adrien Gaidon.
\newblock 3d packing for self-supervised monocular depth estimation.
\newblock In {\em CVPR}, 2020.

\bibitem{hazirbas2017fusenet}
Caner Hazirbas, Lingni Ma, Csaba Domokos, and Daniel Cremers.
\newblock Fusenet: Incorporating depth into semantic segmentation via fusion-based cnn architecture.
\newblock In {\em ACCV}. Springer, 2017.

\bibitem{he2016deep}
Kaiming He, Xiangyu Zhang, Shaoqing Ren, and Jian Sun.
\newblock Deep residual learning for image recognition.
\newblock In {\em CVPR}, 2016.

\bibitem{he2022ra}
Mu He, Le Hui, Yikai Bian, Jian Ren, Jin Xie, and Jian Yang.
\newblock Ra-depth: Resolution adaptive self-supervised monocular depth estimation.
\newblock In {\em ECCV}. Springer, 2022.

\bibitem{kusupati2020normal}
Uday Kusupati, Shuo Cheng, Rui Chen, and Hao Su.
\newblock Normal assisted stereo depth estimation.
\newblock In {\em CVPR}, 2020.

\bibitem{lee2021learning}
Seokju Lee, Sunghoon Im, Stephen Lin, and In~So Kweon.
\newblock Learning monocular depth in dynamic scenes via instance-aware projection consistency.
\newblock In {\em AAAI}, volume~35, 2021.

\bibitem{lee2021attentive}
Seokju Lee, Francois Rameau, Fei Pan, and In~So Kweon.
\newblock Attentive and contrastive learning for joint depth and motion field estimation.
\newblock In {\em CVPR}, 2021.

\bibitem{li2023bevdepth}
Yinhao Li, Zheng Ge, Guanyi Yu, Jinrong Yang, Zengran Wang, Yukang Shi, Jianjian Sun, and Zeming Li.
\newblock Bevdepth: Acquisition of reliable depth for multi-view 3d object detection.
\newblock In {\em AAAI}, volume~37, 2023.

\bibitem{long2021adaptive}
Xiaoxiao Long, Cheng Lin, Lingjie Liu, Wei Li, Christian Theobalt, Ruigang Yang, and Wenping Wang.
\newblock Adaptive surface normal constraint for depth estimation.
\newblock In {\em ICCV}, 2021.

\bibitem{lu2023deep}
Guoyu Lu.
\newblock Deep unsupervised visual odometry via bundle adjusted pose graph optimization.
\newblock In {\em ICRA}, pages 6131--6137, 2023.

\bibitem{indoor}
Guoyu Lu, Yan Yan, Li Ren, Philip Saponaro, Nicu Sebe, and Chandra Kambhamettu.
\newblock Where am i in the dark: Exploring active transfer learning on the use of indoor localization based on thermal imaging.
\newblock {\em Neurocomputing}, 173:83--92, 2016.

\bibitem{localizeiccv}
Guoyu Lu, Yan Yan, Li Ren, Jingkuan Song, Nicu Sebe, and Chandra Kambhamettu.
\newblock Localize me anywhere, anytime: a multi-task point-retrieval approach.
\newblock In {\em ICCV}, pages 2434--2442, 2015.

\bibitem{lyu2021hr}
Xiaoyang Lyu, Liang Liu, Mengmeng Wang, Xin Kong, Lina Liu, Yong Liu, Xinxin Chen, and Yi Yuan.
\newblock Hr-depth: High resolution self-supervised monocular depth estimation.
\newblock In {\em AAAI}, volume~35, 2021.

\bibitem{mahjourian2018unsupervised}
Reza Mahjourian, Martin Wicke, and Anelia Angelova.
\newblock Unsupervised learning of depth and ego-motion from monocular video using 3d geometric constraints.
\newblock In {\em CVPR}, 2018.

\bibitem{pilzer2018unsupervised}
Andrea Pilzer, Dan Xu, Mihai Puscas, Elisa Ricci, and Nicu Sebe.
\newblock Unsupervised adversarial depth estimation using cycled generative networks.
\newblock In {\em 3DV}. IEEE, 2018.

\bibitem{scharstein2002taxonomy}
Daniel Scharstein and Richard Szeliski.
\newblock A taxonomy and evaluation of dense two-frame stereo correspondence algorithms.
\newblock {\em International journal of computer vision}, 47, 2002.

\bibitem{tang2022perception}
Yang Tang, Chaoqiang Zhao, Jianrui Wang, Chongzhen Zhang, Qiyu Sun, Wei~Xing Zheng, Wenli Du, Feng Qian, and Juergen Kurths.
\newblock Perception and navigation in autonomous systems in the era of learning: A survey.
\newblock {\em TNNLS}, 2022.

\bibitem{telea2004image}
Alexandru Telea.
\newblock An image inpainting technique based on the fast marching method.
\newblock {\em Journal of graphics tools}, 9, 2004.

\bibitem{wang2018learning}
Chaoyang Wang, Jos{\'e}~Miguel Buenaposada, Rui Zhu, and Simon Lucey.
\newblock Learning depth from monocular videos using direct methods.
\newblock In {\em CVPR}, 2018.

\bibitem{wang2023planedepth}
Ruoyu Wang, Zehao Yu, and Shenghua Gao.
\newblock Planedepth: Self-supervised depth estimation via orthogonal planes.
\newblock In {\em CVPR}, 2023.

\bibitem{wang2023sqldepth}
Youhong Wang, Yunji Liang, Hao Xu, Shaohui Jiao, and Hongkai Yu.
\newblock Sqldepth: Generalizable self-supervised fine-structured monocular depth estimation.
\newblock {\em arXiv preprint arXiv:2309.00526}, 2023.

\bibitem{watson2021temporal}
Jamie Watson, Oisin Mac~Aodha, Victor Prisacariu, Gabriel Brostow, and Michael Firman.
\newblock The temporal opportunist: Self-supervised multi-frame monocular depth.
\newblock In {\em CVPR}, 2021.

\bibitem{yan2021channel}
Jiaxing Yan, Hong Zhao, Penghui Bu, and YuSheng Jin.
\newblock Channel-wise attention-based network for self-supervised monocular depth estimation.
\newblock In {\em 3DV}. IEEE, 2021.

\bibitem{zhang2023lite}
Ning Zhang, Francesco Nex, George Vosselman, and Norman Kerle.
\newblock Lite-mono: A lightweight cnn and transformer architecture for self-supervised monocular depth estimation.
\newblock In {\em CVPR}, 2023.

\bibitem{zhao2022monovit}
Chaoqiang Zhao, Youmin Zhang, Matteo Poggi, Fabio Tosi, Xianda Guo, Zheng Zhu, Guan Huang, Yang Tang, and Stefano Mattoccia.
\newblock Monovit: Self-supervised monocular depth estimation with a vision transformer.
\newblock In {\em 3DV}. IEEE, 2022.

\bibitem{zhou2017unsupervised}
Tinghui Zhou, Matthew Brown, Noah Snavely, and David~G Lowe.
\newblock Unsupervised learning of depth and ego-motion from video.
\newblock In {\em CVPR}, 2017.

\end{thebibliography}
}

\end{document}